\documentclass[10pt,twocolumn,letterpaper]{article}

\pdfoutput=1 
\usepackage{iccv}
\usepackage{times}
\usepackage{epsfig}
\usepackage{graphicx}
\usepackage{amsmath}
\usepackage{amssymb}

\usepackage{comment}
\usepackage{times}
\usepackage{soul}
\usepackage[utf8]{inputenc}
\usepackage{caption}
\usepackage{booktabs}
\usepackage{color}
\usepackage{algorithm}
\usepackage{algorithmic}
\usepackage{array}
\usepackage{multirow}
\usepackage{enumitem}
\usepackage{verbatim}
\usepackage{subcaption}
\iccvfinalcopy % *** Uncomment this line for the final submission

\def\iccvPaperID{5} % *** Enter the ICCV Paper ID here

\ificcvfinal\fi

\newcommand{\norm}[1]{\left\lVert#1\right\rVert}
\newcommand{\rpm}{\raisebox{.2ex}{$\scriptstyle\pm$}}
\begin{document}

\title{ProtoGAN: Towards Few Shot Learning for Action Recognition}

\author{Sai Kumar Dwivedi\\
Mercedes-Benz R\&D India\\
{\tt\small saikumar.dwivedi@daimler.com}
\and
Vikram Gupta\\
Mercedes-Benz R\&D India\\
{\tt\small vikram.gupta@daimler.com}
\and
Rahul Mitra\\
IIT Bombay\\
{\tt\small rmitter@cse.iitb.ac.in}
\and
Shuaib Ahmed\\
Mercedes-Benz R\&D India\\
{\tt\small shuaib.ahmed@daimler.com}
\and
Arjun Jain\\
IIT Bombay, Axogyan AI\\
{\tt\small arjunjain@gmail.com}
}
\maketitle
\ificcvfinal\thispagestyle{empty}\fi

%%%%%%%%% ABSTRACT
\begin{abstract}
Few-shot learning (FSL) for action recognition is a challenging task of recognizing novel action categories which are represented by few instances in the training data. In a more generalized FSL setting (G-FSL), both seen as well as novel action categories need to be recognized. Conventional classifiers suffer due to inadequate data in FSL setting and inherent bias towards seen action categories in G-FSL setting. In this paper, we address this problem by proposing a novel ProtoGAN framework which synthesizes additional examples for novel categories by conditioning a conditional generative adversarial network with class prototype vectors. These class prototype vectors are learnt using a Class Prototype Transfer Network (CPTN) from examples of seen categories. Our synthesized examples for a novel class are semantically similar to real examples belonging to that class and is used to train a model exhibiting better generalization towards novel classes. We support our claim by performing extensive experiments on three datasets: UCF101, HMDB51 and Olympic-Sports. To the best of our knowledge, we are the first to report the results for G-FSL  and provide a strong benchmark for future research. We also outperform the state-of-the-art method in FSL for all the aforementioned datasets.
\end{abstract}

%%%%%%%%% BODY TEXT
\section{Introduction}

Action recognition has been a long-standing and actively pursued problem in the computer vision community due to its practical applications in areas such as surveillance, semantic video retrieval and multimedia mining. Recently, Convolutional Neural Network (CNN) based methods~\cite{i3d,c3d,temporal_seg_action} have achieved tremendous success in recognizing actions from videos in supervised learning paradigm. However, the performance of these methods deteriorates drastically~\cite{dense_dilated} when recognizing action classes that are not adequately represented in their training data (\emph{novel} classes). This  problem limits the deployment of these methods in real world applications where the number of action classes to be recognized increases rapidly with use cases. In certain cases, the number of samples for novel classes are quite few for even traditional data augmentation techniques~\cite{upsampling} to work. Therefore, systems with more advanced learning paradigm like \emph{Few Shot Learning}~\cite{low-shot}, which learns to recognize novel classes from only a few examples (\emph{few-shots}), have come into prominence. 

Few-shot learning problem can be classified into two broad settings~\cite{low-shot} based on their evaluation protocols: standard FSL (\emph{FSL}) and the more realistic Generalized-FSL (\emph{G-FSL}). FSL focuses on recognition of only \emph{novel} classes during evaluation whereas under G-FSL, a combination of both \emph{seen} (adequately represented at training) and \emph{novel} classes is considered. The presence of \emph{seen} classes during evaluation can incorrectly bias a classifier to predict a \emph{seen} class when the input belongs to a \emph{novel} class. Hence, the G-FSL setting is considered more challenging. There have been several approaches~\cite{metagan,prototypical,low-shot} to tackle FSL for image classification. The most notable among them use \emph{meta-learning}, \emph{representation learning} and \emph{generative modelling}. Meta-learning mimics the few-shot inference time scenario during training, representation learning tries to learn the similarity of new samples to existing few-shots and generative modelling augments the novel classes with synthetic data. However, similar approaches for action recognition remain quite under-explored.
% A recent and popular solution is to augment the training data with synthesized examples~\cite{m%etagan,dagan} through generative methods. 

\begin{figure*}[t]
    \centering
    %\captionsetup{width=\textwidth}
    \includegraphics[width=\textwidth]{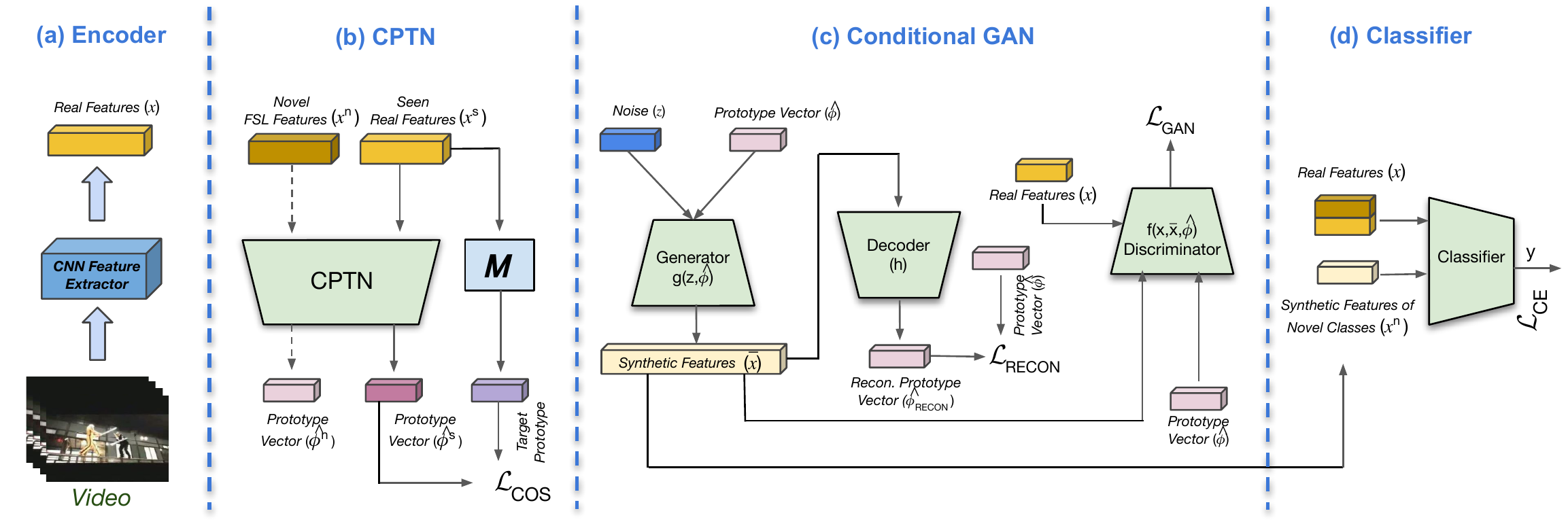}
    \caption{Illustration of our proposed \emph{ProtoGAN} framework. (a) Encoder extracts video features using spatio-temporal CNN (b) \emph{Class Prototype} vector $\phi^n$ for a novel class is learnt by mapping video features of seen data $x^n$ in CPTN to target formed through feature aggregator function $\mathcal{M}$ using cosine similarity loss $L_{COS}$. The solid line denotes the training time path and dotted line denotes the inference path. (c) Learnt class prototype vector is used as a conditioning element in CGAN to synthetically generate more samples using generator $g$ and discriminator $f$ using adversarial loss $L_{GAN}$. Decoder $h$ is used to reconstruct the prototype vector from generator's output using $L_{RECON}$ to ensure discriminative properties. (d) Classifier takes real seen features, few-shot features and synthetic features of novel classes to train with $L_{CE}$ loss}
    \label{fsl_pipeline}
\end{figure*}

Generative methods like Conditional GAN (CGAN)~\cite{CGAN} synthesizes additional data for \emph{novel} classes using a conditioning element. In image classification, Antoniou et al. \cite{dagan} explored data augmentation GAN which uses the few-shot samples directly as a conditioning element to generate synthetic data. Zhang et al. \cite{metagan} use statistics from the samples as an alternative for the conditioning element. However, when the number of shots is less, conditioning element in these methods does not represent the class semantics. 

To this end, we propose the \emph{ProtoGAN} framework which conditions CGAN on a \emph{class-prototype} vector to synthesize additional video features for the action recognition classifier. Class-prototype vector is \emph{learnt} through a feature aggregator network called \emph{Class Prototype Transfer Network} (CPTN). The synthetic features generated using our learned conditioning element for a class are semantically similar to actual features of videos belonging to that class. We justify our claims by performing extensive experiments on three publicly available datasets namely \emph{UCF101}~\cite{ucf101}, \emph{HMDB}~\cite{hmdb} and \emph{Olympic-Sports}~\cite{olympic} under G-FSL and FSL settings. We focus the necessary ablations in our framework on G-FSL as it presents a more practical setting applicable for real world scenarios. The key contributions of our paper can be summarized as follows:
\begin{enumerate}
  \item Introducing a \emph{learned class-prototype} vector for videos capturing class level semantics which is subsequently used as conditioning element for a CGAN to generate relevant synthetic features for novel classes.
  \item To the best of our knowledge, we are the first to demonstrate results for Generalized-Few Shot Learning (G-FSL) on publicly available action recognition datasets and hence provide a strong benchmark for future research.
 \item Outperform state-of-the-art method under the standard FSL setting for the three aforementioned datasets and across different number of samples or shots.
\end{enumerate}

%The most notable among them use \emph{meta-learning}, \emph{representation learning} and \emph{generative modelling}. Meta-learning method learns how to learn from limited labelled data through training by training sub-tasks called \emph{episodes} which mimics the FSL test time scenario. Representation learning projects data into lower dimension embedding for mapping to novel classes. In generative based approaches, additional examples are synthesized by training generative adversarial network (GAN)~\cite{CGAN}.
%Unlike previous approaches~\cite{compund_memory,WACV}, we evaluate our framework for both seen as well as novel classes. We are the first to report results of action recognition under G-FSL setting by conducting comprehensive experiments on public datasets. The efficacy of the proposed framework is shown by outperforming state-of-the-art results on FSL based action recognition. The usefulness of the transformation network is established through ablation studies. 

\section{Related Work}
FSL approaches in image classification literature are pre-dominantly based on meta-learning ~\cite{metalearning_semisupervised,metagan,model_Agnostic}. Under the umbrella of meta-learning, various approaches like metric learning~\cite{prototypical,vinyals2016matching,Relation_network}, learning optimization~\cite{optimization} and learning to initialize and finetune~\cite{model_Agnostic} are proposed. Metric learning methods learn the similarity between images~\cite{vinyals2016matching} to classify samples of novel classes at inference based on nearest neighbors with labelled samples. Learning optimization involves memory based networks like LSTM~\cite{fsl_machine_web} or memory augmented networks~\cite{SNAIL} which aim to replace stochastic gradient descent optimizer with the help of external memory. Learning to initialize and fine-tune aims to make minimal changes in the network when adapting to new task of classifying novel classes with few examples. Although, all the works mentioned above attempt to solve the broad problem of few-shot learning, many of them do not consider the full potential of seen classes as they learn the models in k-shot n-way method~\cite{SNAIL}. %Moreover, meta-learning methods are not memory efficient~\cite{prototypical}.
In~\cite{low-shot}, the authors propose a method of representation learning of visual features from the seen classes and translate it to novel classes for images. Recently, GAN based synthesizing of novel classes is gaining importance~\cite{dagan} where the novel class is augmented with synthetic features. In~\cite{metagan}, a generic framework to augment data with conditioning GAN on sample mean class-representative vector is explored. A detailed overview of existing methods of FSL in images is given in~\cite{fsl_survey}.

In contrary to FSL in images, video-based FSL methods such as action recognition have got less attention. In~\cite{compund_memory}, authors have proposed a compound memory architecture which transforms a variable length video sequence into a fixed length matrix helping in few shot classification. A method for action localization in FSL setting is explored in~\cite{1shot_sequence}. Attribute-based feature generation for unseen classes from GAN by using Fisher vector representation was explored in zero-shot learning in~\cite{zsl_ijcai2018}. Authors in~\cite{WACV} used Gaussian based generative approach to augment data for novel classes, where each action is represented in the visual space as a probability distribution. Since, the experimental setup in~\cite{WACV} includes classifying all novel classes together in contrary to k-shot n-way setting, this forms a strong baseline to compare with our proposed framework in FSL setting.

\textbf{Our approach:} To avoid classifier's bias towards seen classes, unlike the approaches mentioned above, we propose learning a prototype vector which captures the semantics of a class by using features from seen classes. A CGAN is trained to learn the mapping from prototype vector to visual features. Additional data for novel classes is synthesized by using the predicted prototype vector from CPTN as conditioning element in CGAN.

\section{Proposed Method}
\label{sec:proposed-method}
An overview of our proposed framework can be seen in Fig.~\ref{fsl_pipeline} which can be broadly segregated into four major blocks: (a) \emph{Encoder} for extracting video features (b) \emph{Class-Prototype Transfer Network} for creating class prototype vector, (c) \emph{CGAN} for generating synthetic features when conditioned on class-prototype vector, and (d) \emph{Classifier} for predicting correct action class.

%% Move this to related work section
%%The usage of a C-WGAN in context of Zero-Shot action recognition has been presented in~\cite{zsl_ijcai2018}. In contrast, our method provides a novel formulation of learning a concise representation of a class which is used as the conditioning element in C-WGAN. Our class representative element is a low dimension embedding learnt from raw video features of that particular class.  Our learned representation summarizes the intra-class semantics better than other heuristic dimension reduction methods like average or max pool.  More sophisticated dimension reduction scheme such as PCA is not applicable for \emph{novel} classes due to lack of sufficient examples. We then train a classifier on the newly augmented training set for action recognition. A schematic view of our learning framework is shown in Fig.~\ref{}. Broadly our approach can be segregated into three 3 stages as follows.

\subsection{Preliminaries}
Let $x$, $\bar{x}$, $y$, $\hat{\phi}$ represent real video features, synthetic video features, class labels and class prototype vectors, respectively. Let $\mathcal{S}$ = $\{x^s$, $y^s | x^s \in \mathcal{X}^s$, $y^s \in \mathcal{Y}^s\}$ be the training set for seen classes where $x^s \subset \mathbb{R}^{d_x}$ denotes the spatio-temporal features, $y^s$ denotes the class labels in $\mathcal{Y}^s = \{y^{s}_1,\ldots,y^{s}_S\}$ with $S$ seen classes. The class prototype vector $\hat{\phi}^s \in \Phi^s \subset \mathbb{R}^{d_\phi}$ is calculated from $x^s$ video features. Let the cardinality of each member in $\mathcal{Y}^s$ be denoted by $\mathcal{K}^s = \{k_1^s,\ldots,k_S^s\}$. Additionally, $\mathcal{N} =\{x^n, y^n) | {x^n} \in \mathcal{X}^n, y^n \in \mathcal{Y}^n\}$ is available during training from novel categories where $y^n$ is a class from a disjoint label set $\mathcal{Y}^n = \{y^{n}_1,\ldots,y^{n}_N\}$ of $N$ novel labels. Similarly, the cardinality of each member in $\mathcal{Y}^n$ is denoted by $\mathcal{K}^n = \{k^n,\ldots,k^n\}$ where $k^n \ll min(K^s)$ and $min()$ is defined as minimum value in the set. The class prototype vector of novel class $\hat{\phi}^n$ is inferred through CPTN. In GFSL, the task is to learn a classifier $Q_{gfsl} : \mathcal{X}^s \cup \mathcal{X}^n \rightarrow \mathcal{Y}^s \cup \mathcal{Y}^n $ and in FSL the task is to learn a classifier $Q_{fsl}: \mathcal{X}^n \rightarrow \mathcal{Y}^n$.

\subsection{Class Prototype Transfer Network (CPTN)\label{approach_cptn}}
Generation of synthetic samples for a class using a CGAN~\cite{CGAN} requires a conditioning element capturing the semantics of that class. When the number of samples belonging to the class is high, statistical methods like \emph{mean} of the samples tend to accurately represent those semantics~\cite{class_mean}. However, in the case of \emph{novel} classes with few examples, this doesn't hold true as such methods are susceptible to capture noisy intricacies due to lack of sufficient data. To this end, we learn a mapping function \emph{Class-Prototype Transfer Network} (CPTN) through \emph{feature aggregation} where features of each video belonging to a particular class is mapped to a lower dimension embedding serving as the representative of that class. We term this vector embedding as the \emph{class-prototype} vector $\hat{\phi}$. We use seen classes data containing large number of samples to model this mapping and then transfer the mapping to novel categories.

%\vikram{The property of the prototype vector: represent class and drop intricacies} \sai{Have mentioned below by the choice of design. Do you want explicit mention?}

%The network is \emph{learnt} by regressing the output from $\phi_{i}^s$ to ${\phi^s}_{C}$, where $C$ is the total number of classes to the target prototype vector.We propose a \emph{Class-Prototype Transfer Network} (CPTN) to model the mapping function $\mathcal{M}: x^s \rightarrow {\phi^s}$ through . 

Feature aggregation $\mathcal{M}$ is a two step process where the sample representation $\mathcal{R}$ is followed by dimensionality reduction $\mathcal{T}$ as per the following equation:
\begin{equation}
    \label{eq:cptn}
    \begin{gathered}
        \mathcal{M}_{x^s \rightarrow {\hat{\phi}^s}} \:=\: \mathcal{T}\left( \mathcal{R}(x^s)\right); \quad
        \mathcal{R}(x^s) \:=\: \frac{1}{n}  \sum_{i = 1}^{n} x_{ci}^s  %\quad\quad
        %\mathcal{T} \:=\:  \mathcal{P}(\mathcal{R}(x^s))
    \end{gathered}
\end{equation}

where $n$ is the number of instances of class $c$, $\mathcal{R}$ is the mean of the samples, $\mathcal{T}$ is a spatial dimensionality reduction function like average/max pool which is applied on the output of $\mathcal{R}$ and $\hat{\phi}^s$ is the prototype vector of seen class formed through process $\mathcal{M}$. We have used average-pool in our approach. The feature aggregation step ensures that meaningful information is retained while getting rid of intricate and noisy details specific to individual videos. Cosine similarity loss $L_{COS}$ is used as a loss function for training the CPTN and is given by the equation: 
\begin{equation}
    \label{eq:cptn_loss}
    \mathcal{L}_{COS} \:=\: \frac{{{\phi^s}}}{\norm{{{\phi^s}}}} \: \dot \: \frac{{\hat{\phi}^s}}{\norm{{\hat{\phi}^s}}}
\end{equation}

where, ${\phi^s}$ is the predicted class-prototype vector and $\hat{\phi}^s$ is the target formed by the Eq.~\ref{eq:cptn} for seen classes. Once the network is trained, the \emph{prototype} vector ${\hat{\phi}^n}_{c}$  for a sample video belonging to a novel class $c$ is obtained by passing its input video features $x^n_c$  through the CPTN. For subsequent stages, we use $\hat{\phi}^n$ and $\hat{\phi}^s$ as class prototype vectors for novel and seen classes, respectively. The training framework is shown in (b) block of Fig.~\ref{fsl_pipeline}.

\begin{table*}[t]
\begin{subtable}{\textwidth}
\scalebox{0.97}{
\begin{tabular}{l|ccc|ccc|ccc|}
\multicolumn{10}{c}{UCF101} \\
\toprule
 & \multicolumn{3}{c}{1-shot} & \multicolumn{3}{|c|}{3-shot} & \multicolumn{3}{c|}{5-shot} \\ 
 \cmidrule{2-10}
 & \textit{S} & \textit{N} & \textit{H} & \textit{S} & \textit{N} & \textit{H} & \textit{S} & \textit{N} & \textit{H} \\ 
  \cmidrule{1-10}
 \emph{Base-Classifier} 
    & 82.7\rpm0.6& 38.4\rpm2.0& 52.4\rpm1.9
    & 83.1\rpm5.4& 61.2\rpm3.0& 70.3\rpm1.2
    & 88.2\rpm0.9& 68.8\rpm1.9& 77.3\rpm0.9 \\
\emph{Heuristic-Proto} 
    & 88.0\rpm1.0& 46.3\rpm2.2& 60.7\rpm2.0
    & 92.0\rpm0.7& 62.1\rpm1.9& 74.1\rpm1.2
    & 94.0\rpm1.4& 68.7\rpm1.5& 79.3\rpm0.7 \\
\emph{Sample-Proto} 
    & 88.0\rpm1.9& 45.9\rpm2.4& 60.3\rpm2.2
    & 92.0\rpm0.9& 61.9\rpm1.9& 74.0\rpm1.3
    & 94.4\rpm0.8& 68.4\rpm1.3& 79.3\rpm0.8\\
\emph{Learned-Proto} 
    & 75.3\rpm1.3& 52.3\rpm2.2& \textbf{61.7}\rpm \textbf{1.6}
    & 87.7\rpm0.8& 64.9\rpm1.7& \textbf{74.6}\rpm\textbf{1.0}
    & 90.5\rpm0.9& 71.3\rpm1.2& \textbf{79.7}\rpm\textbf{0.8} \\ 
\bottomrule
\end{tabular}{}
}
\end{subtable}{}

\begin{subtable}{\textwidth}
\scalebox{0.97}{
\begin{tabular}{l|ccc|ccc|ccc| }
\multicolumn{10}{c}{HMDB} \\
\toprule
 &\multicolumn{3}{c}{1-shot} & \multicolumn{3}{|c|}{3-shot} & \multicolumn{3}{c|}{5-shot} \\ 
 \cmidrule{2-10}
 Methods & \textit{S} & \textit{N} & \textit{H} & \textit{S} & \textit{N} & \textit{H} & \textit{S} & \textit{N} & \textit{H} \\ 
\cmidrule{1-10}
\emph{Base-Classifier} 
    & 59.9\rpm1.3& 12.7\rpm2.4& 20.9\rpm3.3
    & 52.5\rpm1.1& 35.7\rpm2.0& 42.4\rpm1.4
    & 61.4\rpm4.6& 38.7\rpm3.0& 47.2\rpm1.0 \\
\emph{Heuristic-Proto} 
    & 53.3\rpm0.9& 19.5\rpm1.6& 28.5\rpm1.8 
    & 59.7\rpm2.4& 35.4\rpm2.1& 44.3\rpm1.3
    & 57.9\rpm1.9& 44.5\rpm1.4& 50.3\rpm0.6 \\
\emph{Sample-Proto} 
    & 53.4\rpm1.8& 19.7\rpm1.6& 28.7\rpm1.6
    & 59.6\rpm2.1& 35.3\rpm2.0& 44.3\rpm1.3
    & 58.8\rpm3.5& 43.8\rpm1.7& 50.1\rpm0.7\\
\emph{Learned-Proto} 
    & 51.9\rpm1.5& 25.8\rpm1.4& \textbf{34.4}\rpm\textbf{1.3}
    & 58.9\rpm1.8& 37.4\rpm1.4& \textbf{45.7}\rpm\textbf{0.9}
    & 61.6\rpm3.0& 43.3\rpm1.5& \textbf{50.9}\rpm\textbf{0.6} \\ 
\bottomrule
\end{tabular}
}
\end{subtable}{}

\begin{subtable}{\textwidth}
\scalebox{0.97}{
\begin{tabular}{l|ccc|ccc|ccc| }
\multicolumn{10}{c}{Olympic}\\
\toprule
 &\multicolumn{3}{c}{1-shot} & \multicolumn{3}{|c|}{3-shot} & \multicolumn{3}{c|}{5-shot}\\ 
 \cmidrule{2-10}
 Methods & \textit{S} & \textit{N} & \textit{H} & \textit{S} & \textit{N} & \textit{H} & \textit{S} & \textit{N} & \textit{H}\\ 
  \cmidrule{1-10}
\emph{Base-Classifier} 
    & 94.5\rpm3.1& 16.9\rpm3.1& 28.6\rpm4.4
    & 93.2\rpm3.2& 41.1\rpm2.9& 56.9\rpm2.9
    & 92.0\rpm3.3& 54.9\rpm3.2& 68.7\rpm2.0 \\
\emph{Heuristic-Proto} 
    & 95.5\rpm3.2& 18.5\rpm2.9& 30.9\rpm4.0
    & 95.3\rpm3.5& 41.2\rpm2.8& 57.5\rpm2.6
    & 95.0\rpm3.7& 54.8\rpm3.2& 69.4\rpm2.1 \\
\emph{Sample-Proto} 
    & 95.5\rpm3.7& 18.6\rpm2.9& 31.0\rpm4.8
    & 95.1\rpm3.4& 41.9\rpm2.8& 58.1\rpm2.6
    & 94.8\rpm3.9& 55.7\rpm3.4& 70.0\rpm2.3 \\
\emph{Learned-Proto} 
    & 94.8\rpm3.4& 20.9\rpm2.6& \textbf{34.1}\rpm\textbf{4.7}
    & 93.5\rpm3.0& 46.0\rpm2.7& \textbf{61.5}\rpm\textbf{2.2}
    & 93.2\rpm3.4& 59.2\rpm3.4& \textbf{72.2}\rpm\textbf{2.0} \\ 
\bottomrule
\end{tabular}{}
}
\end{subtable}{}
\caption{Comparison of Action Recognition accuracy (\%) of our proposed framework with other baselines on UCF101, HMDB and Olympic datasets under G-FSL settings. \emph{Base-Classifier} denotes vanilla classifier with standard augmentation. \emph{Heuristic-Proto} and \emph{Sample-Proto} denotes proposed baselines inspired from prototype-vectors used in image classification. \textit{S}, \textit{N}, \textit{H} represents \emph{seen}, \emph{novel} and \emph{harmonic mean}, respectively. We report mean accuracy from 20 different training runs and standard deviation is reported by $\pm$.}
\label{table:GFSL}
\end{table*}

\subsection{Conditional GAN \label{approach_gan}}
This module in our \emph{ProtoGAN} framework involves a CGAN~\cite{CGAN} to generate synthetic video features ($\bar{x}^{n}$) for novel classes by using the conditioning element ${\hat{\phi}^n}$ learned in the previous CPTN stage. The Wasserstein loss~\cite{wgan} is chosen over vanilla GAN loss as it provides more stability in training in low-data regime~\cite{zsl_ijcai2018}. The Wasserstein loss in a CGAN between real features $p_{r}$ and synthetically generated feature $p_{s}$ is given by,

\begin{equation}
            \label{eq:cwgan}
             \begin{split}
    \mathcal{L}_{WGAN} = \mathbb{E}_{x \sim p_{r}} \left[ f(x, {\hat{\phi}}) \right] \:-\: \mathbb{E}_{z \sim p_z} \left[f(g(\bar{x}), \hat{\phi})) \right] \:-\: \\
    \alpha\mathbb{E}[(||\nabla_{\bar{x}}f(\hat{x},\hat{\phi})||_2 - 1)^2] \: ;
            \quad \left(\bar{x} \:=\: g({\hat{\phi}}, z)\right) \sim p_{s}
            \end{split}
\end{equation}

where $x$'s are \emph{real} video features drawn from $p_{r}$. $\hat{x}$ is a convex combination of $x$ and $\bar{x}$, $\hat{\phi}$ is the conditioning element of a particular class, ($f$) is the discriminator, ($g$) is the generator, ${z} \sim \mathbf{N}(0,I)$ is the noise vector and $\alpha$ is the penalty coefficient. The first two terms in Equation.~\ref{eq:cwgan} approximate the Wasserstein distance and the third term is the penalty for constraining the gradient of ($f$) to have unit norm along the convex combination of real and generated pairs. The \emph{Class-prototype} vector embedding  ${\hat{\phi}}$ along with random noise $z$ are sent as input the generator ($g$). It generates an output with a dimension same as that of input video features. The generated video features $\bar{x}$ along with real video features $x$ are passed to the discriminator ($f$) which is trained with an adversarial loss as per Eq.~\ref{eq:cwgan}. Hence at equilibrium, the generator produces video features which are similar to the real features. 

The generated features of a particular class $y_i$ should be similar to the real features of the that class and farther away from the features of other classes. As it is compute intensive to find the closest match for a generated feature from a set of real features of that class, generated and real features are grouped to form unmatched pairs (for different classes). Cosine embedding loss $\mathcal{L}_{EMD} = \max(0,cos(x_i,\bar{x}_j))$ (for $j\neq i$) is used to compute the distance for unmatched pairs.

%To prevent mode collapse during generation of synthetic features \cite{multimodal_modecollapse} and 
To ensure that the generated features contain class semantics for subsequent supervised classification, similar to~\cite{multimodal_cycle}, an decoder ($h$) is used. It generates $\phi_{RECON}$ as an output in an attempt to reconstruct back $\hat\phi$ from $g_{\theta}(\hat\phi,z)$. We use cosine similarity loss $L_{RECON}$ for the reconstruction. Thus, the net loss of CGAN is given by, 
\begin{equation}
\label{eq:stage2-tot-loss}
\begin{split}
 \mathcal{L}_{total} \:=\: \min_{g} \max_{f}\mathcal{L}_{WGAN} + \lambda\mathcal{L}_{RECON} + \gamma\mathcal{L}_{EMD} 
\end{split}
\end{equation}

where $\lambda$ is the hyper-parameter for weighting the reconstruction loss and $\gamma$ for embedding loss. Training framework of the current stage is shown in (c) block of Fig.~\ref{fsl_pipeline}. 

%Each cluster represents a $\phi^n_c$'s from a different class.  
\begin{table*}[t]
    \centering
        \begin{tabular}{l |cc| cc| cc}
        \toprule
        & \multicolumn{2}{c|}{UCF101} & \multicolumn{2}{c|}{HMDB} & \multicolumn{2}{c}{Olympic}\\
        \cmidrule{1-7}
        Method & 1-shot & 5-shot & 1-shot & 5-Shot & 1-shot & 5-shot \\
        \cmidrule{1-7}
        \emph{Sample-Proto} & 0.248 & 0.117 & 0.341 & 0.205 & 0.297 & 0.193 \\
        \emph{Learned-Proto}  & 0.219 & 0.106 & 0.323 & 0.198 & 0.272 & 0.191 \\
        \bottomrule
        \end{tabular}
    \caption{Cosine distance between mean of synthetic features and mean of test data for novel classes. Results are reported for all the datasets in 1-shot and 5-shot setting. Lower number is better.}
\label{tab:cdist}
\end{table*}

\begin{table*}[]
    \centering
        \begin{tabular}{l |cc| cc}
        \toprule
        & \multicolumn{2}{c|}{UCF101} & \multicolumn{2}{c}{HMDB} \\
        \cmidrule{1-5}
        $\mathcal{T} \:\ Function$ & 1-shot & 5-shot & 1-shot & 5-Shot \\
        \cmidrule{1-5}
        \emph{No-Pool}      & 57.8\rpm1.7 & 79.8\rpm1.1 
                            & 23.3\rpm2.3 & 51.2\rpm0.8 \\
        \emph{Max-Pool}     & 60.6\rpm1.3 & 79.6\rpm0.9
                            & 24.3\rpm2.5 & 50.7\rpm0.8 \\
        \emph{Average-Pool} & 61.7\rpm1.6 & 79.7\rpm0.8
                            & 34.4\rpm1.3 & 50.9\rpm0.6 \\
        \bottomrule
        \end{tabular}
    \caption{Comparison of Action Recognition accuracy (\%) of our proposed framework with different dimensionality reduction function $\mathcal{T}$ on UCF101 and HMDB datasets under G-FSL settings. \emph{No-Pool} denotes use of no dimensionality reduction function, \emph{Max-Pool} denotes max-pooling and \emph{Average-Pool} denotes average-pooling. We report harmonic mean accuracy from 20 different training runs and standard deviation is reported by $\rpm$.}
    \label{table:cptn_ablation}
\end{table*}

\subsection{Classifier \label{approach_classifier}}
We utilize the generator $(g)$ learned in the previous stage to produce additional video features for \emph{novel} classes given their corresponding \emph{class-prototype} ${\phi_c^n}$ obtained through the CPTN network. The classifier is trained in a supervised learning paradigm with the real features of seen and novel classes and synthetic features of novel classes. We subsequently train action recognition classifiers $Q_{gfsl}$ and $Q_{fsl}$ for G-FSL and FSL settings, respectively, with Cross-Entropy loss $\mathcal{L}_{CE}$. 

We made an insightful observation that generated samples having high reconstruction error ($\mathcal{L}_{RECON}$) are not suitable for training the classifier as they are semantically different than its corresponding class. Hence, we use a pruning method to remove all the synthetic features with high reconstruction loss. Features with low reconstruction loss tend to be more discriminative. This is because the decoder tries to reconstruct the class-prototype vector which is the representative of the class.

%However, all the generated synthetic features are not suitable for training the classifier as they do not maintain class semantics. Thus, we use a pruning method to remove all the synthetic features with high reconstruction loss. Features with low reconstruction loss tend to be more discriminative. This is because the decoder tries to reconstruct the class-prototype vector which is the representative of the class \sai{check arjun's comments}.

%Training set of $Q_{gfsl}$ on the training set consisting of \emph{real} video features ($x^n$) augmented with the additional synthetic examples($\bar{x}^{n}$). 
%Formally, the classifier is denoted by $Q : \psi \cup {\psi}^{\prime} \rightarrow \mathcal{Y}^s \cup \mathcal{Y}^u$ 

\section{Experimental Setup} \label{sec:experiment}
Details about the experimental setup are discussed in this section. Implementation details such as network architecture and training procedures are explained in~Section.~\ref{implementation}, dataset used for evaluation are described in Section.~\ref{dataset} and evaluation protocol followed for experiments under G-FSL and FSL setting are highlighted in Section.~\ref{sec:eval_protocol}.

\begin{figure*}[t]
    \centering
    %\captionsetup{width=\textwidth}
    \includegraphics[trim=50 30 10 10,clip,width=0.8\textwidth,height=0.4\textwidth]{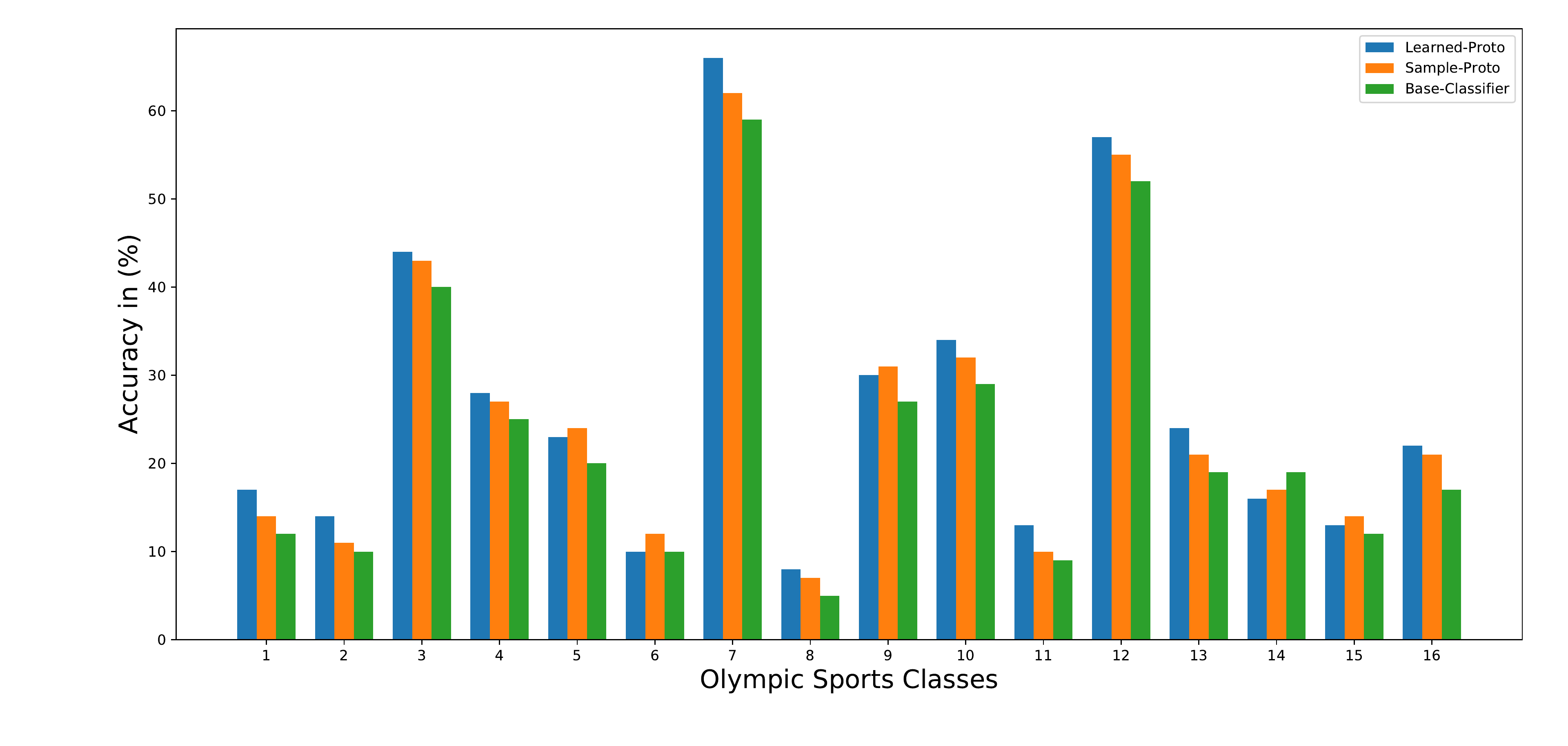}
    \caption{Mean classification accuracy (in \%) of our proposed \emph{ProtoGAN} framework with other baselines for all classes (when present as novel classes in different experimental runs) for olympic sports dataset. Classes are numbered according to alphabetical order of names. We report mean accuracy from 10 different runs. Best viewed in color.}
    \label{per_cls_acc}
\end{figure*}

\begin{table*}[h]
    \centering
        \begin{tabular}{l |cc| cc}
        \toprule
        & \multicolumn{2}{c|}{UCF101} & \multicolumn{2}{c}{HMDB} \\
        \cmidrule{1-5}
        Method & 1-shot & 5-shot & 1-shot & 5-Shot \\
        \cmidrule{1-5}
        \emph{Without-Pruning}  & 60.8\rpm1.9 & 79.5\rpm0.9 
                                & 23.7\rpm2.6 & 50.7\rpm0.8 \\
        \emph{With-Pruning}     & 61.7\rpm1.6 & 79.7\rpm0.8 
                                & 34.4\rpm1.3 & 50.9\rpm0.6 \\
        \bottomrule
        \end{tabular}
    \caption{Comparison of action recognition accuracy (\%) of our proposed framework with and without pruning the synthetic features under G-FSL setting. We report harmonic mean accuracy from 20 different training runs and standard deviation is reported by $\rpm$.}
    \label{table:classifier_ablation}
\end{table*}

\subsection{Implementation Details \label{implementation}}
\begin{itemize}
    \item \textbf{\textit{Encoder}} -
    %\subsubsection{Encoder}
    We extracted video features from \emph{C3D} network~\cite{c3d} pre-trained on Sports-1M dataset~\cite{sports_million} where each video was divided into  non-overlapping chunks of 16 frames and the mean of the \emph{fc6} layer outputs in~\cite{c3d}, of size 4096, was taken.

    \item \textbf{\textit{Class-Prototype Transfer Network (CPTN)}} - 
    %\subsubsection{Class-Prototype Transfer Network (CPTN)}
    A two-layer MLP was used with an input size of 4096, a hidden layer of 1024 and output layer (\emph{class-prototype} vector embedding $\hat{\phi}$) of 128 neurons with \emph{Sigmoid} as activation function. The model was trained with cosine similarity loss with ADAM~\cite{adam} optimizer initialized with a learning rate of 0.005 and weight-decay of 0.0005. The training was run for 50 epochs.

    %%%%%%%%%%%%%%%%%%%%%%%%%%%%%%%%%%%%%%%%%%%%%%%%%%%%%%%%%%%%%%%%%%%%%
    % Move it to experiment folder
    %Further, in the paper, this formulation is referred to as \emph{Learned-Proto} for comparison purpose. Additionally, we studied a baseline to understand the impact of our CPTN network. In the baseline formulation, \emph{class-prototype} embedding for all videos belonging to a class is calculated by taking sample mean. This mean was further used as a conditioning element to the CGAN for synthetic feature generation. We term this formulation as \emph{Heuris-Proto}.

    \item \textbf{\textit{CGAN}} -
    %\subsubsection{CGAN and Classifier}
    The generator $(g)$ is a two layer fully connected network with input size of $256$ ($128 (\hat{\phi_c})$ + $128 (noise)$) and output size of $4096$. The decoder $(h)$ used in cyclic-reconstruction is a two-layer fully connected with input and output size being 4096 and 128, respectively. The discriminator $(f)$ is a two layer fully connected neural network with input dimension of 4096 and output dimension of 1 deciding whether the features are real or fake. Learning rate of 0.001 and with a weight decay of 0.0001 was used to train the CGAN for 25 epochs using ADAM optimizer. \emph{LeakyReLU} was used as activation function for $(g)$, $(f)$ and $(h)$. We set $\lambda$ and $\gamma$ in Eq.~\ref{eq:stage2-tot-loss} to be $0.01$ and $0.1$, respectively. The gradient penalty parameter $\alpha$ for WGAN is set to 10.

    \item \textbf{\textit{Classifier}} -
    The final action recognition classifier is a single layer MLP with input as $4096$ and output as the number of classes. Cross-entropy loss was used with ADAM optimizer with a learning rate of $0.001$. For each novel class, we generated twice the maximum number of samples present in a seen class, sort them in ascending order of their reconstruction loss and pick the top $50\%$.
\end{itemize}

\subsection{Evaluation Protocols \label{sec:eval_protocol}}
To highlight the efficacy of our proposed framework, we thoroughly evaluate our method under G-FSL and FSL setting for all the action recognition datasets. A brief description of the evaluation protocol are as follows:
\begin{itemize}[noitemsep]
    \item \textbf{\textit{G-FSL setting}} - In this setting, the test set consists of \emph{seen} $S$ and \emph{novel} $N$ classes and the model is evaluated based on their classification accuracy. The accuracy of seen, novel and harmonic mean \emph{H} as per Eq.~\ref{eq:harmonic} of the two are reported.
    \begin{equation}
            \label{eq:harmonic}
            Acc_{h} \:=\:  \frac{2.Acc_{s}.Acc_{n}}{Acc_{s} + Acc_{n}}
    \end{equation}
    where $Acc_s$ and $Acc_n$ are accuracy of \emph{seen} and \emph{novel} classes, respectively.
    
    \item \textbf{\textit{FSL setting}} - In this setting, the test split consists of only \emph{novel} classes $N$.
\end{itemize}

In our experiments, under $k$-shot setting, $k$ random samples were chosen from a set of available examples for each \emph{novel} class. To eliminate bias towards any given set of samples and (\emph{seen}, \emph{novel}) class split, we repeat our experiments 20 times, with randomly chosen splits and samples. We report the mean results. For instance, in the 3-shot learning setting, for each of the 20 training runs over different class splits, 3 random samples were drawn for each \emph{novel} class.

%we compare our proposed framework against existing approaches under the standard FSL setting and provide a strong benchmark when evaluation is done for G-FSL due to lack of existing approaches.0

\begin{table*}[t]
\centering
    \begin{tabular}{p{1.5cm} cccccc}
    \toprule
     & & \multicolumn{5}{c}{k-shots} \\ 
     \cmidrule{3-7}
     & Methods & 1 & 2 & 3 & 4 & 5 \\ 
    \midrule
    
    \multirow{2}{*}{UCF101} & \cite{WACV}  & -  & 68.7\rpm3.3 & 73.5\rpm2.2 & 76.5\rpm2.1 & 78.6 \rpm1.8\\
    & \emph{Learned-Proto}  & 57.8\rpm3.0 & \textbf{71.1}\rpm\textbf{3.0} & \textbf{75.3}\rpm\textbf{2.7} & \textbf{78.0}\rpm\textbf{1.8} & \textbf{80.2}\rpm\textbf{1.3}\\
    
    \midrule
    
    \multirow{2}{*}{HMDB} &\cite{WACV}  & -  & 42.1\rpm3.6& 47.5\rpm3.3& 50.3\rpm3.4& 52.5\rpm3.1 \\
     & \emph{Learned-Proto}  & 34.7\rpm9.2 & \textbf{43.8}\rpm\textbf{5.4} & \textbf{49.1}\rpm\textbf{5.1} & \textbf{52.3}\rpm\textbf{4.2} & \textbf{54.0}\rpm\textbf{3.9}  \\
    
    \midrule
    
    \multirow{2}{*}{Olympic} &\cite{WACV} & -  & 73.2\rpm7.4& 75.3\rpm7.3& 80.2\rpm7.2& 83.8\rpm7.1 \\
     & \emph{Learned-Proto}  & 71.6\rpm9.4 & \textbf{75.0}\rpm\textbf{7.4} & \textbf{78.4}\rpm\textbf{6.2} & \textbf{82.1}\rpm\textbf{5.6} & \textbf{86.3}\rpm\textbf{5.1} \\

    \bottomrule
    \end{tabular}
\caption{Comparison of action recognition accuracy (in \%) of our proposed \emph{ProtoGAN} framework against current state-of-the-art on UF101, HMDB and Olympic dataset under FSL settings. We report mean accuracy from 20 different training runs and standard deviation is reported by $\pm$. Recognition accuracy for 1-shot is missing for \cite{WACV}.}
%Recognition accuracy for 1-shot is missing for \cite{WACV}. $LP_{Ours}$ represents Learned-Proto}
\label{table:FSL}
\end{table*}

\subsection{Datasets \label{dataset}}
We evaluate our method on three publicly available datasets for action recognition,

\begin{itemize}[noitemsep]
    \item UCF101~\cite{ucf101}: contains 13320 videos spanned over 101 classes. For our experiments we randomly split the data into 51 \emph{seen} and 50 \emph{novel} classes.
    \item HMDB51~\cite{hmdb}: contains 6766 videos spanned over 51 classes. For our experiments we randomly split the data into 26 \emph{seen} and 25 \emph{novel} classes.
    \item Olympic Sports~\cite{olympic}: contains 783 videos spanned over 16 classes. We refer Olympic Sports as Olympic in subsequent sections. For our experiments we randomly split the data into 8 \emph{seen} and 8 \emph{novel} classes. 
\end{itemize}

\begin{comment}
\begin{figure}[t]
    \centering
    \includegraphics[trim=10 10 10 10,clip,width=0.49\textwidth,height=0.35\textwidth]{images/per_cls_acc.eps}
    \caption{TSNE plots of synthetic features generated from CGAN for 20 randomly choosen classes from UCF101 dataset. Each color represents a different class where (a) Features obtained when GAN is conditioned with \emph{Heuristic-Proto} and (b) Features obtained when GAN is conditioned on \emph{Learned-Proto}. Red dotted circles represent clusters.} 
    \label{fig_tsne_hmdb}
\end{figure}
\end{comment}

\section{Experimental Results}\label{sec:exp-results}
Details about our experimental results are discussed in this section. Results for our proposed framework under G-FSL setting are compared with different baselines in Section. \ref{sec:gfsl}. Section. \ref{sec:ablation} describes ablation studies highlighting the importance of various components in our framework. Comparison with state-of-the art method under FSL setting is discussed in Section. ~\ref{sec:fsl}.

\subsection{Generalized-Few Shot Learning (G-FSL)}\label{sec:gfsl}
% \subsubsection{Baseline Comparison}
As the results under G-FSL on the aforementioned action recognition datasets are reported for the first time, we designed the baseline methods by taking inspiration from the \emph{Few-Shot Learning} methods proposed in image classification tasks. Similar to~\cite{metagan}, sample mean of the examples from a class was taken as its class-prototype and we term this method as \emph{Heuristic-Proto} in our experiments. In \emph{Sample-Proto} the video features were directly taken as the class-prototype as described in~\cite{dagan}. Our proposed class prototype vectors generated through CPTN are termed as \emph{Learned-Proto} in the experiments. To evaluate the performance of the entire framework, a vanilla classifier - \emph{Base-Classifier} was directly trained on \emph{C3D} video features with standard augmentation mentioned in~\cite{upsampling}. The results for all the above approaches on UCF101, HMDB and Olympic datasets are reported in Table.~\ref{table:GFSL}. 

As can be seen in Table.~\ref{table:GFSL}, \emph{Heuristic-Proto}, \emph{Sample-Proto} and \emph{Learned-Proto} outperform the \emph{Base-Classifier} with a huge margin for all the shots in all the datasets. This establishes that, the addition of generated features removes classifier's bias towards \emph{seen} classes which is depicted by the gain of accuracy for novel classes. Taking mean of the samples for novel classes, decreases the performance of \emph{Heuristic-Proto} for 1-shot while giving competitive results for 5-shots. This can be attributed to poor representation of the conditional element by taking mean of single example. \emph{Sample-Proto} performs better than other two baselines but shows higher standard deviation. This is because its performance depends on the quality of chosen samples as the class prototype is represented by the samples themselves without any aggregation. %Hence, the choice of \emph{few-shots} significantly affects the performance. 
In contrast, transferred statistics through CPTN in our proposed \emph{ProtoGAN} framework reduces this adverse effect. \emph{Learned-Proto} outperforms baseline methods by a maximum of 1.4\%, 5.7\% and 2.9\% on UCF101, HMDB and Olympics, respectively, for the 1-shot setting.
%While \emph{Sample-Proto} gives comparatively better results than other baselines but sees a high standard deviation.

%Adding synthetic features gives a gain of more than 10\% on UCF101 and HMDB in the 1-shot setting. The gain decreases with an increase in the number of shots and in 5-shot it is close to 2\% in UCF101 and 4\% in HMDB. However, in the case of Olympic sports, the gain (4\%) is lower due to a smaller dataset size. For CPTN and CGAN to give a maximum performance boost, seen classes should be enough samples.

Figure. \ref{per_cls_acc} illustrates the comparison of per class mean accuracy of our proposed \emph{ProtoGAN} framework with other baselines. Mean accuracy is taken for all classes when present as novel classes during 10 different experimental runs. As can be seen from the figure, our proposed framework performs better in 11 out of 16 classes in Olympic Sports dataset in the range of 1-4\%. However, the margin of inaccuracies is less significant as compared to the accurate classes. This re-establishes the superiority of the \emph{ProtoGAN} framework.

\subsection{Ablation Studies}\label{sec:ablation}

\subsubsection{Quality of Synthetic Features}
To verify and compare the quality of the synthesized features of \emph{novel} classes using different conditional elements, we try to 
quantify their similarity with real video features available from the test set. Specifically, we take mean of all the synthesized and real features for a \emph{novel} class separately and compute cosine distance between them. The mean cosine distance over all the \emph{novel} classes for 1 and 5-shots are reported in Table.~\ref{tab:cdist}. Results are reported for UCF101, HMDB and Olympic datasets. One can observe, examples generated using our \emph{Learned-Proto} for 1-shot matches the distribution of real features much more accurately than that of \emph{Sample-Proto}.

\subsubsection{Dimensionality Reduction}
Dimensionality reduction function ($\mathcal{T}$) plays a crucial role in getting rid of intricate details and preserving class semantics. To demonstrate the effect of ($\mathcal{T}$), a model was trained without applying $\mathcal{T}$ and another with Max-Pool instead of Average-Pool. As can be in seen in Table.~\ref{table:cptn_ablation}, the recognition accuracy drops significantly when no ($\mathcal{T}$) is used and thus restating that dropping intricacies helps in creating better \emph{class-prototype} vector. Max-Pool is slightly inferior in performance compared to Average-Pool. A reason for this is that features obtained after average-pool has more aggregate information than max-pool, hence provides more stable features. 

\subsubsection{Pruning}
To validate the effect of pruning synthetic features in increasing order of their reconstruction loss, the proposed \emph{ProtoGAN} framework was trained without it and the results are reported in Table.~\ref{table:classifier_ablation}. The superior performance of the classifier trained on data after pruning validates the efficacy of the reconstruction loss in creation and subsequent selection of meaningful examples. However, the gain is higher for HMDB as compared to UCF101. This can be attributed to less number of seen examples which affects the quality of class-prototype vectors and hence the generator.

\subsection{Few Shot Learning (FSL)}\label{sec:fsl}
A comparison of our \emph{ProtoGAN} framework 
against the current state-of-the-art approach~\cite{WACV} under FSL setting is presented in Table.~\ref{table:FSL}. 
The authors of ~\cite{WACV} have also used \emph{C3D}~\cite{c3d} video features in their evaluation. Our method outperforms \cite{WACV} in all k-shots with similar or lower standard deviation. Note that the results for 1-shot are not reported in~\cite{WACV}. The improved performance can be attributed to the usage of a \emph{learned} prototype vector. The \emph{learned} vector computed through a non-linear function via a network provides a better alternative to a vector formed by a linear combination of basis vectors as mentioned in \cite{WACV}.   This establishes the wide applicability of our approach in both G-FSL and conventional FSL settings.
% This establishes that besides providing a strong benchmark in the generalized setting, the proposed framework outperforms state-of-the-art by a significant margin.
%The average gain over all the shots for UCF101, HMDB, and Olympic are 15.5\%, 12.8\%, and 14.0\%, respectively with \emph{I3D} features.

%{\sai why only one method}
%Table.~\ref{table:FSL}, we report the accuracy obtained by our proposed method \emph{ProtoGAN} along with the current state-of-the-art approach~\cite{WACV}. We observe our method outperforms in all shots. The results for 1-shot are not reported by the state-of-art approach. Our method outperforms by a margin of 14.1\%, 13.1\% and 10.3\% in 5-shot setting for UCF101, HMDB and Olympic datasets, respectively. The average gain over all the shots except for 1-shot are 15.5\%, 12.8\% and 14.0\%, respectively.

\section{Conclusion}
\label{sec:conclusion}
In this paper, we present a novel \emph{ProtoGAN} framework which synthesizes video features for novel categories by conditioning a CGAN with a \emph{class-prototype} vector embedding to address the problem of \emph{Few-Shot Learning} for action recognition. Class-prototype vector is learnt through a feature aggregator network called Class Prototype Transfer Network (CPTN). The performance of the proposed framework was evaluated on three publicly available datasets for both seen and novel classes under G-FSL setting for the first time. We obtained encouraging results showing the efficacy of the proposed framework under G-FSL settings on action recognition and established a strong benchmark for future research. Under standard FSL setting, we outperform state-of-the-art method on all the datasets across different shots.
\\
\\
\textbf{Acknowledgement:} We gratefully acknowledge Brijesh Pillai and Partha Bhattacharya at Mercedes-Benz R\&D India, Bangalore
for providing the funding and infrastructure for this work.

{\small
%\bibliographystyle{ieee}
%\bibliography{egbib}

}

\end{document}